\documentclass{article}%
\usepackage{amsmath}
\usepackage{amsfonts}
\usepackage{amssymb}
\usepackage{graphicx}
\usepackage{algorithmic}
\usepackage{algorithm}
\usepackage{hyperref}%
\providecommand{\U}[1]{\protect \rule{.1in}{.1in}}
\providecommand{\U}[1]{\protect \rule{.1in}{.1in}}

\begin{document}

\title{A novel gradient-based method for decision trees optimizing arbitrary
differential loss functions }
\author{Andrei V. Konstantinov and Lev V. Utkin\\Higher School of Artificial Intelligence Technologies\\Peter the Great St.Petersburg Polytechnic University\\St.Petersburg, Russia\\e-mail: andrue.konst@gmail.com, lev.utkin@gmail.com}
\date{}
\maketitle

\begin{abstract}
There are many approaches for training decision trees. This work introduces a
novel gradient-based method for constructing decision trees that optimize
arbitrary differentiable loss functions, overcoming the limitations of
heuristic splitting rules. Unlike traditional approaches that rely on
heuristic splitting rules, the proposed method refines predictions using the
first and second derivatives of the loss function, enabling the optimization
of complex tasks such as classification, regression, and survival analysis. We
demonstrate the method's applicability to classification, regression, and
survival analysis tasks, including those with censored data. Numerical
experiments on both real and synthetic datasets compare the proposed method
with traditional decision tree algorithms, such as CART, Extremely Randomized
Trees, and SurvTree. The implementation of the method is publicly available,
providing a practical tool for researchers and practitioners. This work
advances the field of decision tree-based modeling, offering a more flexible
and accurate approach for handling structured data and complex tasks. By
leveraging gradient-based optimization, the proposed method bridges the gap
between traditional decision trees and modern machine learning techniques,
paving the way for further innovations in interpretable and high-performing models.


\end{abstract}

\section{Introduction}

In the context of the ever-increasing volume and complexity of data in various
application tasks, existing data processing and modeling methods do not always
allow for effective forecasting, especially in the presence of censored
tabular data \cite{chen2024introduction} and sequential data processing
\cite{dietterich2002machine}. One of the effective models for tabular data
processing are ensembles of decision trees \cite{wu2008top,ZH-Zhou-2012};
however, they are less flexible than neural networks, which limits their
applicability. More flexible and accurate decision tree-based models are
needed, along with noise- and outlier-resistant training algorithms.
Therefore, new methods for constructing decision trees based on gradient-based
search algorithms for node splitting parameters should be applied. These
methods enable higher prediction accuracy, the construction of trees with
arbitrary differentiable loss functions, and the effective combination of
decision trees with neural networks.

When processing numerical data arrays, decision tree-based models such as
random forests, extremely randomized trees, and gradient boosting of decision
trees are often used. The main advantages of decision tree-based models
include: higher accuracy and lower susceptibility to overfitting compared to
other models, such as neural networks; high efficiency of training algorithms
in terms of performance; interpretability of such models (both individual
trees and ensembles using methods like TreeSHAP
\cite{bifet2022linear,muschalik2024beyond}). Due to these features, decision
trees form the basis for the most accurate models processing tabular data and
represent a promising research direction.

The classic method for constructing decision trees is CART (Classification and
Regression Trees) \cite{Breiman-etal-1984}, which has two versions for
classification and regression tasks. It is based on greedy recursive splitting
of tree nodes using a heuristic rule depending on a task type. Splitting is
performed by iterating over threshold values and evaluating a certain
criterion. For example, in the case of classification, the criterion is the
reduction of impurity in the subtrees. Among all considered splits, the best
one is selected based on the criterion. To determine predictions, a new
example is passed through the tree, and it is assigned the average value of
the target variable of the training examples in the leaf it falls into. In
other words, the leaves of the constructed tree contain the averaged values of
the training observations. In the case of classification, class frequencies
are used as such values. Note that the tree structure is suboptimal since the
construction is performed greedily, but this drawback is not critical for
practical applications. However, in addition to this, less accurate models may
also result from the discrepancy between the heuristic rule and the loss
function that actually needs to be optimized. When solving real-world tasks
different from standard ones, such as regression and classification, the
defining component of the training algorithm is the loss function. For
example, predicting tasks with censored data cannot be solved without using
specific loss functions, of which there is a wide variety, for examaple, the
C-index or the Brier score \cite{Wang-Li-Reddy-2019}. However, in classical
decision tree construction algorithms, instead of directly optimizing the loss
function, the node splitting rules are determined heuristically. The following
drawbacks of this approach can be highlighted:

\begin{itemize}
\item The critical impact of the algorithm's greediness on increasing model
overfitting, as the model's predictions are determined based on the average
value in the leaf, which, at great depth, leads to minimal error on the
training set and increases generalization error.

\item The inability to adapt to new loss functions, especially when they
depend on external factors (e.g., when the loss function is determined by a
neural network).
\end{itemize}

To build an accurate model based on optimizing the loss function using
decision trees, gradient boosting can be applied. The idea of boosting is to
iteratively construct a set of trees, where each subsequent tree is built to
refine the model. In the case of gradient boosting, each tree corresponds to a
gradient descent step. Note that this method can optimize any differentiable
loss function using only its derivative. This allows training such models
based on trees as part of a unified model with neural networks, where training
can only be performed using gradient methods. However, the decision trees in
such a model are built based on heuristic splitting rules rather than the
gradient of the loss function. Importantly, in all existing boosting models,
including XGBoost \cite{Chen-Guestrin-2016}, trees are built based on
once-calculated values of the first (and in some cases, second) derivatives,
which do not change during the tree construction process, leading to the
aforementioned drawbacks. In other words, the trees themselves are built to
optimize the function using its derivatives.

These limitations become particularly evident in complex tasks such as
survival analysis, where specific loss functions (e.g., log-likelihood for
censored data) are required. To address these challenges, this work introduces
a novel gradient-based approach for constructing decision trees, enabling the
optimization of \emph{arbitrary differentiable loss functions} and
facilitating the integration of decision trees with neural networks. The
approach does not require a specific splitting criterion and a function
determining leaf values. Instead of averaging the target variable values, as
in classical algorithms, at each step of the proposed method, predictions are
refined using the first and second derivatives of the loss function, which are
calculated based on the training examples corresponding to the given node.

The approach not only improves prediction accuracy but also enables the use of
complex loss functions, such as those required for survival analysis or tasks
involving censored data. Furthermore, the method allows us to combine decision
trees with neural networks, creating hybrid models capable of handling
structured data (e.g., sequences, graphs) while maintaining the
interpretability of tree-based models.

The contributions of this work can be summarized as follows:

\begin{enumerate}
\item A novel gradient-based method for constructing decision trees that
optimizes arbitrary differentiable loss functions is proposed.

\item It is shown how the proposed method can be applied to solving specific
tasks, including the regression and classification tasks, the survival
analysis task.

\item Numercial experiments on real and synthetic datasets illustrate
outperformance of the method in classification, regression, and survival
analysis tasks.
\end{enumerate}

The code implementing the proposed method is publicly available at: \url{https://github.com/NTAILab/gradient_growing_trees}.

The paper is organized as follows. Related work considering the existing
explanation methods can be found in Section 2. A formal problem statement for
constructing the decision trees which optimize a loss function is considered
in Section 3. The method for constructing decision trees with arbitrary loss
functions is provided in Section 4. Application of the proposed method to
classification and regression tasks as well as survival analysis are
considered in Section 5. Numerical experiments comparing the different types
of decision trees are given in Section 6. Concluding remarks are provided in
Section 7. Appendix contains some peculiarities of the proposed method
software implementation.

\section{Related work}

One of the modifications of traditional decision trees is the oblique decision
trees \cite{murthy1994system,Wickramarachchi-etal-16}, which can be viewed as
an extension of traditional decision trees that use oblique (multivariate)
splits instead of axis-parallel univariate splits. While traditional decision
trees split the feature space using hyperplanes parallel to the axes, oblique
trees use hyperplanes at an angle, allowing for more complex and flexible
decision boundaries. Weighted oblique trees were proposed in
\cite{yang2019weighted} where different weights assign to each instance for
child nodes at all internal nodes by soving an optimization problem with
differentiable objective function.

Another modification is the use of the support vector machine for each
decision in the tree \cite{bennett1998support}. An extension of traditional
decision trees to decision trees based on imprecise probabilities for handling
noisy data was presented in \cite{Mantas-Abellan-2014}.

Multivariate Decision Trees which consider multiple features simultaneously
for splitting, improving their ability to model interactions between features,
were studied in \cite{boutilier2023optimal,brodley1995multivariate}. Another
modification, aimed at combining symbolic decision trees with approximate
reasoning offered by fuzzy representation, was proposed in
\cite{janikow1998fuzzy}. Fuzzy rule-based decision trees employing non-linear
splits by representing them as fuzzy conjunction rules were introduced in
\cite{wang2015fuzzy}. Optimal decision trees for binary variables were
presented in \cite{hu2019optimal}.

Survival decision trees have been designed specifically for survival analysis,
which involves predicting the time until an event of interest occurs
\cite{Bou-Hamad-etal-2011,Gordon-Olshen-1985,Hu-Steingrimsson-2018,Segal-1988}%
. Survival trees with time-dependent covariates were studied in
\cite{Huang-Chen-Soong-1998}. An Optimal Survival Trees algorithm using
mixed-integer optimization and local search techniques to generate globally
optimized survival tree models was proposed in \cite{bertsimas2022optimal}.

Many types of decision trees and their detailed study are considered in
surveys
\cite{costa2023recent,de2013decision,kotsiantis2013decision,rokach2005decision,sharma2016survey}%
.

\section{Decision Trees Optimizing a Loss Function}

Let a twice-differentiable loss function $l(y,\hat{y})$ be given, where $y$ is
the label of the training example, and $\hat{y}$ is the current model
prediction for this training example. Note that the labels and predictions can
be from different spaces; for example, the class label can represent the class
number, and the prediction can be a vector of probabilities for all classes.
Let also a training dataset $D=\{(x_{i},y_{i})\}_{i=1}^{N}$ be given from the
joint distribution $(X,Y)\sim P$, where each feature vector $x_{i}$ consists
of $d$ components $(x_{i}^{(1)},\dots,x_{i}^{(d)})\in \mathbb{R}^{d}$, and
$y_{i}$ is the corresponding label determining the loss for the $i$-th
example. The goal is to construct a decision tree $f(x)$ minimizing the
empirical risk functional:
\begin{equation}
\hat{L}[f]=\frac{1}{N}\sum_{i=1}^{N}l(y_{i},f(x_{i}))\approx \mathbb{E}%
_{X,Y\sim P}[l(Y,f(X))],
\end{equation}
while having high generalization ability. To ensure high generalization
ability, it is proposed to minimize:
\begin{equation}
L[f]=\hat{L}[f]+\Omega \lbrack f],
\end{equation}
where $\Omega$ is a regularization penalty component, and the desired function
$f$ belongs to the class of functions $F$ corresponding to decision trees with
given hyperparameters, such as maximum depth, minimum number of examples in a
node for splitting, and so on.

Let at the current iteration of the algorithm, a tree $f\in F$ be constructed,
containing leaves $L$, and each leaf $n\in L$ is assigned a leaf prediction
$c_{n}$, which is the tree's prediction for any input feature vector $x$
falling into the region of space corresponding to the leaf, denoted further as
$R_{n}$. Note that the union of all such regions gives $\mathbb{R}^{d}$, and
all regions are pairwise disjoint. Then the decision tree $f$ can be
represented as:
\begin{equation}
f(x)=\sum_{n\in L}I[x\in R_{n}]\cdot c_{n},
\end{equation}
where $I$ is the indicator function.

The correctness of this expression follows from the fact that the leaves are
pairwise disjoint, so only the term corresponding to the leaf into which the
example falls will be non-zero. To determine the condition $[x\in R_{n}]$, it
is sufficient to check that the example falls into the parent node and that
from the parent node, the example falls into the given node, which is
determined by checking the predicate $x^{(k_{p})}\leq \theta_{p}$, where $p$ is
the parent node, $\theta_{p}$ is the threshold value in the node, and $k_{n}$
is the feature index for splitting. If the predicate is true, the example
falls into the left subtree; otherwise, it falls into the right subtree.

At each step of tree construction, one of the leaf nodes is split into two
nodes, generating two new regions depending on the threshold value and the
feature index for splitting. Thus, to construct a tree aiming to minimize the
loss function, it is necessary to determine the rule for splitting node $n $
and the rule for refining the prediction values in the leaves (child nodes).

\section{Constructing Decision Trees with Arbitrary Loss Functions}

Consider a leaf node $n$. Before its splitting, for all points from the region
$R_{n}$ (in the leaf), the corresponding prediction function of the tree is
constant and equal to the current leaf value, i.e., the following holds:
\begin{equation}
\forall x\in R_{n}\ (f(x)=c_{n}).
\end{equation}

For splitting, only the training examples falling into the region $R_{n}$ are
considered. To obtain a split, it is proposed to represent the desired tree
after splitting as the sum of the current prediction function in the leaf and
a new step function $\phi_{n}$, refining the values in the left and right
subtrees:
\begin{equation}
\phi_{n}(x)=I[x^{(k_{n})}\leq \theta_{n}]\cdot u_{n}+I[x^{(k_{n})}>\theta
_{n}]\cdot v_{n},
\end{equation}
where $u_{n},v_{n}$ are the adjusted values for the left and right
predictions, respectively, i.e., they are left and right values of the
predicted target variable.

The key idea of the proposed method is that the new function, for all points
falling into the node, is defined as the sum of the previous value in the node
and the adjusting function:
\begin{equation}
\tilde{f}(x)=(c_{n}+\phi_{n}(x)).
\end{equation}

The splitting parameters are determined to minimize the error of the new node
function $\tilde{f}(x)$. To do this, first, the loss function is expanded into
a series around the previous unchanged estimate of the target variable value
in the leaf $c_{n}$, for $x\in R_{n}$:
\begin{align}
l(y,\tilde{f}_{n}(x))  &  =l(y,c_{n})+\phi_{n}(x)\cdot \left.  \frac{\partial
l(y,z)}{\partial z}\right \vert _{z=f(x)}\nonumber \\
&  +\frac{1}{2}\phi_{n}^{2}(x)\cdot \left.  \frac{\partial^{2}l(y,z)}{\partial
z^{2}}\right \vert _{z=f(x)}+o(\phi_{n}^{2}(x)).
\end{align}

For brevity, denote the first and second derivatives as:
\begin{equation}
g_{n}(x,y)=\left.  \frac{\partial l(y,z)}{\partial z}\right \vert _{z=f_{n}%
(x)},\ h_{n}(x,y)=\left.  \frac{\partial^{2}l(y,z)}{\partial z^{2}}\right \vert
_{z=f_{n}(x)}.
\end{equation}

Then a new problem of minimizing the loss function with the regularization
$\Omega$ under the condition of validity of the series expansion, taking into
account that all considered examples fall into node $n$, is:
\begin{align}
L_{n}  &  =\frac{1}{N}\sum_{i=1}^{N}I[x_{i}\in R_{n}]\cdot \lbrack \phi
_{n}(x_{i})\cdot g_{n}(x_{i},y_{i})\nonumber \\
&  +\frac{1}{2}\phi_{n}^{2}(x_{i})\cdot h_{n}(x_{i},y_{i})]+\Omega(u_{n}%
,v_{n};k_{n},\theta_{n}).
\end{align}

Introduce the following notations for brevity:
\begin{equation}
U_{i}=I[x_{i}^{(k_{n})}\leq \theta_{n}];\ V_{i}=I[x_{i}^{(k_{n})}>\theta
_{n}];\ I_{i}=I[x_{i}\in R_{n}].
\end{equation}

Then the loss function can be split into terms:
\begin{equation}
L_{n}=L_{n}^{U}+L_{n}^{V}+\Omega,
\end{equation}
where
\begin{equation}
L_{n}^{U}=\sum_{i=1}^{N}I_{i}\cdot \lbrack U_{i}(u_{n}\cdot g_{n}(x_{i}%
,y_{i})+\frac{1}{2}u_{n}^{2}\cdot h_{n}(x_{i},y_{i}))],
\end{equation}
and $L_{n}^{V}$ is defined similarly. Let
\begin{equation}
G_{n}^{U}=\sum_{i=1}^{N}I_{i}\cdot U_{i}\cdot g_{n}(x_{i},y_{i}),\ H_{n}%
^{U}=\sum_{i=1}^{N}I_{i}\cdot U_{i}\cdot h_{n}(x_{i},y_{i}),
\end{equation}%
\begin{equation}
G_{n}^{V}=\sum_{i=1}^{N}I_{i}\cdot V_{i}\cdot g_{n}(x_{i},y_{i}),\ H_{n}%
^{V}=\sum_{i=1}^{N}I_{i}\cdot V_{i}\cdot h_{n}(x_{i},y_{i}).
\end{equation}

It is also worth noting that to ensure the validity of replacing the original
loss function with its approximation, it is necessary to control the absolute
value of the adjustment $\phi_{n}$. The first option for ensuring boundedness
is to introduce a learning rate $0<\gamma \leq1$, in which case the final
adjustment function will look as follows:
\begin{equation}
\tilde{\phi}_{n}(x)=\gamma \cdot \phi_{n}(x).
\end{equation}

However, a more correct way to introduce a constraint on $\phi_{n}$ is to set
$l_{2}$-regularization. It is proposed to set the regularization as:
\begin{equation}
\Omega=\frac{M}{2}\lambda \Vert(u_{n},v_{n})^{T}\Vert_{2}^{2}, \label{regul_1}%
\end{equation}
where $M=\sum_{i=1}^{N}I[x_{i}\in R_{n}]$ is the number of points falling into
the leaf; $\lambda$ is a hyper-parameter which controls the strength of the regularization.

Since $\Omega$ can also be split into two terms depending separately on
$u_{n}$ and $v_{n}$, the corresponding components of the loss function can be
minimized separately, similarly to the work \cite{Konstantinov-Utkin-23}:
\begin{equation}
u_{n}=-\frac{G_{n}^{U}}{M\cdot \lambda+H_{n}^{U}},\ v_{n}=-\frac{G_{n}^{V}%
}{M\cdot \lambda+H_{n}^{V}}.
\end{equation}

Thus, for fixed splitting parameters $\theta_{n},k_{n}$, determining
$U_{i},V_{i}$, the optimal values $u_{n},v_{n}$ can be found.

Next, it is proposed to perform partial or complete enumeration of features
and splitting thresholds, similarly to classical algorithms such as CART,
calculate the optimal parameters, and evaluate each potential split by
substituting the obtained values into the loss function approximation $L_{n} $.

The prediction value in the node is determined using the first and second
derivatives. To split the node, the procedure for calculating the first and
second derivatives for all examples falling into the given node is first
called. Then, the total value of the first and second derivatives is
calculated, and the optimal split is searched for. For this, each feature is
considered, the values are sorted by the feature, and for each potential
split, the optimal adjusting values (left and right) are determined, and the
loss function is calculated based on them.

It is extremely important that the choice of the optimal split is performed in
linear time, i.e., the loss function is not evaluated for all examples for
each potential split. This is possible because the partial loss functions
$L_{n}^{U}$ and $L_{n}^{V}$ can be calculated through accumulated sums, which
are also used to calculate the values:
\begin{equation}
L_{n}^{U}=u_{n}\cdot \left(  \sum_{i=1}^{N}I_{i}\cdot g_{n}(x_{i}%
,y_{i})\right)  +u_{n}^{2}\cdot \frac{1}{2}\left(  \sum_{i=1}^{N}I_{i}\cdot
h_{n}(x_{i},y_{i})\right)  .
\end{equation}

Note that during splitting, the value in the leaves depends on the previous
value in the node $c_{n}$, as the gradient of the loss function is calculated
at the point $c_{n}$. In other words, as the tree is constructed, the gradient
of the loss function is recalculated each time a split occurs. Unlike all
other existing algorithms, including those using common gradient values for
all nodes, as in XGBoost \cite{Chen-Guestrin-2016}, in the proposed algorithm,
the loss function is expanded into a series around a more accurate
approximation in the leaf, making the series expansion valid at large learning
rates and allowing the tree to be extended even when the loss function
changes. Thus, it is possible, for example, to combine neural networks with
decision trees, where the neural network acts as the loss function. In this
case, the proposed algorithm does not require calculating the loss function
value for all split candidates, which positively affects performance.

To determine the value in the root node, it is proposed to perform one
gradient descent step, namely, to calculate the first and second derivatives
for all examples in the training dataset. For this, it is necessary to specify
the point at which the loss function expansion is constructed, which can be
interpreted as the initial prediction approximation. The first option is to
use a zero vector, which is natural for both regression and classification
tasks (where zero logits correspond to a uniform discrete probability
distribution). However, in the case of class imbalance or more complex loss
functions, it may be necessary to use another initial approximation calculated
using an auxiliary algorithm, as will be shown later for the classification
task. Therefore, it was also proposed to implement a second option, where the
initial approximation is an input to the decision tree construction algorithm,
along with the hyperparameters.

\section{Application of the Proposed Method to Specific Tasks}

The proposed algorithm is suitable for various machine learning tasks,
including regression, classification, and survival analysis. Additionally,
decision trees constructed for reconstruction tasks can be used for anomaly
detection. Let us consider ways to apply the proposed algorithm for each type
of tasks.

\subsection{Regression task}

For the regression task, the algorithm implements the loss function according
to the classical approach, using the quadratic loss function, where the labels
and predictions are in the same space:
\begin{equation}
l_{SE}(y,\hat{y})=\Vert y-\hat{y}\Vert_{2}^{2}=\sum_{j=1}^{q}(y^{(j)}-\hat
{y}^{(j)})^{2},
\end{equation}
where the first and second partial derivatives with respect to the components
of the prediction vector $\hat{y}$ are, respectively:
\begin{equation}
\frac{\partial l_{SE}(y,\hat{y})}{\partial \hat{y}^{(j)}}=2(\hat{y}%
^{(j)}-y^{(j)}),\  \frac{\partial l_{SE}(y,\hat{y})}{\partial^{2}\hat{y}^{(j)}%
}=2,
\end{equation}
where $j\in \{1,\dots,q\}$, and $q$ is the dimensionality of the target vector.

\subsection{Classification task}

For tasks other than regression, the predictions in the leaves are not in the
same space as the labels and determine, but do not equal, the model's
prediction. To compute the loss function, as well as to determine the final
predictions, an additional transformation is applied to them.

For example, for the classification task, predictions in the leaves determine
\textquotedblleft logits\textquotedblright, and to obtain the final
prediction, the \textquotedblleft softmax\textquotedblright \ operation is
applied to them. For brevity, introduce the following notation:
\begin{equation}
s=(s^{(1)},\dots,s^{(C)})=\text{softmax}(\hat{y}).
\end{equation}

The loss function in this case is a composition of the cross-entropy and the
\textquotedblleft softmax\textquotedblright:
\begin{equation}
l_{CE}(y,\hat{y})=-\sum_{j=1}^{C}I[y=j]\cdot \log(s^{(j)}),
\end{equation}
where $C$ is the number of classes, and the label of the training example
$y\in \{1,\dots,C\}$ determines the class number.

The partial derivatives of this loss function are:
\begin{equation}
\frac{\partial l_{SE}(y,\hat{y})}{\partial \hat{y}^{(j)}}=s^{(j)}-I[y=j],
\end{equation}%
\begin{equation}
\frac{\partial l_{SE}(y,\hat{y})}{\partial^{2}\hat{y}^{(j)}}=s^{(j)}%
\cdot(1-s^{(j)}).
\end{equation}

\subsection{Survival analysis}

To implement nonparametric survival estimates based on the developed decision
tree construction algorithm, a generalized loss function based on the
log-likelihood is proposed. Let a discrete random variable take integer values
from the set $\{1,\dots,C\}$, and let the observation represent not an exact
value but a subset $A\subset \{1,\dots,C\}$ to which the value belongs.
Associate such a subset $A$ with a vector $y$ such that all components at
positions included in the set $A$ are equal to 1, and the remaining components
are equal to 0:
\begin{equation}
y=(I[j\in S])_{j=1}^{C}.
\end{equation}

The loss function is proposed to be set as follows:
\begin{equation}
l_{GCE}(y,\hat{y})=-\log \left(  \sum_{j=1}^{C}y^{(j)}\cdot s^{(j)}\right)
=-\log(y^{T}s),
\end{equation}
which has the following derivatives:
\begin{equation}
\frac{\partial l_{GCE}(y,\hat{y})}{\partial \hat{y}^{(j)}}=s^{(j)}\cdot \left(
1-\frac{y^{(j)}}{y^{T}s}\right)  ,
\end{equation}%
\begin{equation}
\frac{\partial l_{GCE}(y,\hat{y})}{\partial^{2}\hat{y}^{(j)}}=s^{(j)}%
\cdot \left(  1-s^{(j)}-y^{(j)}\cdot \frac{y^{T}s-s^{(j)}}{(y^{T}s)^{2}}\right)
.
\end{equation}

In the case where the subset $A$ consists of one element, the proposed loss
function coincides with cross-entropy. Otherwise, it allows accounting for
imprecise information about the class when the class is not known, but a set
of classes to which the example belongs is known.

Such a generalization allows implementing training for various tasks with
partial labeling, in particular for survival analysis or predictive analytics
tasks with censored data. For this, the elementary outcomes of the modeled
discrete random variable are associated with the event time falling into one
of the non-overlapping time intervals
\begin{equation}
\lbrack \tau_{0},\tau_{1})\cup \lbrack \tau_{1},\tau_{2})\cup \dots \cup \lbrack
\tau_{n},+\infty)=[t_{\min},+\infty),
\end{equation}
where $(n+1)$ is the number of unique event times.

The predictions in the decision tree nodes $\hat{y}$ are logits of the
discrete probability distribution, and the step-wise survival function is
defined as
\begin{equation}
S(t)=1-\sum_{j:\tau_{j}<t}(\text{softmax}(\hat{y}))_{j}.
\end{equation}

The label value of the training example depends on censoring. For an observed
event ($\delta=1$) at time $t$, the label vector is represented as
\begin{equation}
y=(I[t\in \lbrack \tau_{j},\tau_{j+1})])_{j=0}^{n}.
\end{equation}

In the case of a censored observation ($\delta=0$), the label vector is set
as:
\begin{equation}
y=(I[t<\tau_{j}])_{j=0}^{n}.
\end{equation}

Thus, the proposed loss function allows optimizing the likelihood of the
conditional survival function when constructing each decision tree node.

For classification and predictive analytics tasks with censored data, it is
advisable to set the initial approximation based on the estimate of the prior
probability distribution calculated over the entire training dataset. Since
the predictions in the tree nodes are logits, the initial approximation should
also be logits, not a probability distribution. Therefore, it was proposed to
determine the logits based on the estimate of the prior probability
distribution using the following expression:
\begin{equation}
\left(  \text{softmax}\right)  ^{-1}(s)=\{(\ln(s_{j})+\zeta)_{j=1}%
^{C}\ |\  \zeta \in \mathbb{R}\},
\end{equation}
where $\left(  \text{softmax}\right)  ^{-1}(s)$ is the set of admissible
logits (vectors) whose image is the distribution $s$. This expression is
correct because
\begin{align}
\lbrack \text{softmax}((\ln(s_{j})+\zeta)_{j=1}^{C})]_{k}  &  =\frac{\exp
(\ln(s_{k})+\zeta)}{\sum_{j=1}^{C}\exp(\ln(s_{j})+\zeta)}\nonumber \\
&  =\frac{s_{k}\cdot \exp(\zeta)}{\sum_{j=1}^{C}s_{j}\cdot \exp(\zeta)}%
=\frac{s_{k}}{1},
\end{align}
taking into account that $\sum_{j=1}^{C}s_{j}=1$. Also, the described set is
the maximal inclusion set possessing this property, since
\begin{equation}
\ln([\text{softmax}(z)]_{k})=z_{k}-\ln \left(  \sum_{j=1}^{C}\exp
(z_{j})\right)  ,
\end{equation}
i.e.,
\begin{equation}
z_{k}=\ln(s_{k})+\ln \left(  \sum_{j=1}^{C}\exp(z_{j})\right)  ,
\end{equation}
where setting%
\begin{equation}
\zeta=\ln \left(  \sum_{j=1}^{C}\exp(z_{j})\right)  ,\ z_{k}=\ln(s_{k})+\zeta,
\end{equation}
and the condition defining $\zeta$ is an identity, since
\begin{align}
\exp(\zeta)  &  =\sum_{j=1}^{C}\exp(z_{j})=\sum_{j=1}^{C}\exp(\ln(s_{k}%
)+\zeta)\nonumber \\
&  =\sum_{j=1}^{C}s_{k}\cdot \exp(\zeta)=\exp(\zeta).
\end{align}

The choice of the value $\zeta$ can be arbitrary, including setting $\zeta=0$
or $\zeta=-\min \{ \ln(s_{k})\}$, since in all considered loss functions, the
logit values are applied only after the \textquotedblleft
softmax\textquotedblright \ operation, which is invariant to shifting (by a
constant common to all components).

In the case of survival analysis, the prior distribution is determined
according to the Kaplan-Meier model, which defines the probability of an event
occurring after a given time $S(t)$, from which the step distribution function
can be defined as $P(T\leq t)=1-S(t)$. As described above, a discrete
probability distribution over the given time intervals can be considered, in
which case the distribution function becomes the cumulative function of the
discrete distribution, and the interval probabilities are determined as
\begin{equation}
p_{k}=S(\tau_{k})-S(\tau_{k+1}).
\end{equation}

The use of such probability distributions is complicated in cases where zero
components are encountered, since the proposed method for determining logits
requires calculating the logarithm of probabilities. To overcome this problem,
a small value $\epsilon>0$ is introduced, and the components of the artificial
probability distribution are used:
\begin{equation}
\hat{p}_{k}=\frac{\max(p_{k},\epsilon)}{\sum_{j=1}^{C}\max(p_{j},\epsilon)}.
\end{equation}

Then
\begin{align}
\hat{z}_{k}  &  =\ln(\hat{p}_{k})+\zeta=\ln(\max(p_{k},\epsilon))\nonumber \\
&  -\ln \left(  \sum_{j=1}^{C}\max(p_{j},\epsilon)\right)  +\zeta,
\end{align}
which, when setting the constant
\begin{equation}
\zeta=-\ln \left(  \sum_{j=1}^{C}\max(p_{j},\epsilon)\right)  ,
\end{equation}
is equivalent to replacing probabilities less than $\epsilon$ with $\epsilon$
before calculating the inverse transformation.

Another approach to using nonparametric Kaplan-Meier estimates is to calculate
such estimates in the leaves of the trees, similarly to random survival trees
\cite{Bou-Hamad-etal-2011,Gordon-Olshen-1985}. For this, during training,
after completing the tree construction, each leaf is considered, and a
nonparametric survival estimate is calculated. During prediction, for new
feature vectors, the corresponding leaves are determined, and the
nonparametric estimates in the leaves are used as predictions. When using this
approach, the difference from random survival trees lies only in the splitting criterion.

\section{Numerical experiments}

The study of the proposed models and algorithms is conducted using
real\textbf{ }and synthetic data for classification, regression, and survival
analysis. We call the proposed model as \emph{DTLF} (Decision Tree with Loss
Function) for short. Real datasets for classification are Breast cancer, Car,
Dermatology, Diabetic retinopathy, Ecoli, Eeg eyes, Haberman, Ionosphere,
Seeds, Seismic, Soybean, Teaching assistant, Tic-tac-toe, Website phishing,
Wholesale customers. Real datasets for regression are Airfoil, Boston,
Concrete, Diabetes, Wine, Yacht. In addition to real datasets, the following
synthetic datasets are used:

\begin{itemize}
\item \emph{Friedman1} with 5 features generated from a uniform distribution
and expected event time defined as:
\begin{equation}
y=10\sin(\pi x_{1}x_{2})+20(x_{3}-0.5)^{2}+10x_{4}+5x_{5};
\end{equation}

\item \emph{Friedman2} with 4 features from a uniform distribution and
expected event time defined as:
\begin{equation}
y=\sqrt{x_{1}^{2}+(x_{2}x_{3}-1/(x_{2}x_{4}))^{2}};
\end{equation}

\item \emph{Friedman3} with 4 features from a uniform distribution and
expected event time defined as:
\begin{equation}
y=\arctan((x_{2}x_{3}-1/(x_{2}x_{4}))/x_{1});
\end{equation}

\item \emph{Strong Interactions} with 5 features, random interaction
coefficients $w_{ij}$ from a uniform distribution, and expected event time:
\begin{equation}
y=\sum_{i=1}^{d}\sum_{j=i+1}^{d}w_{ij}x_{i}x_{j};
\end{equation}

\item \emph{Sparse Features} with a sparse feature matrix and linear
dependence of expected event time $y = X w$;

\item \emph{Nonlinear Dependencies} with 5 features from a uniform
distribution and nonlinear dependence on each feature, where the expected
event time is defined as:
\begin{equation}
y=4\sin(x_{1})+\log(|x_{2}|+1)+x_{3}^{2}+e^{0.5x_{4}}+\tanh(x_{5});
\end{equation}

\end{itemize}

Event times are generated according to a Weibull distribution with a fixed
shape parameter $k$ and are defined as:
\begin{equation}
T=\frac{y}{\Gamma(1+1/k)}\cdot(-\log(u))^{1/k},
\end{equation}
where $u$ is a random variable uniformly distributed on the interval $[0,1]$,
and the expected event time $y$ depends on the feature vector and is defined
according to the above expressions. For all datasets, the parameter $k=5$.
Censoring labels for all synthetic datasets are randomly generated according
to a Bernoulli distribution with an event probability of $0.8$.

To study the proposed gradient-based decision tree construction algorithm, a
comparison is made between decision trees obtained using the developed
algorithm and those trained using classical algorithms such as CART. We also
use Extremely Randomized Trees (ERT) for comparison purposes. At each node,
the ERT algorithm chooses a split point randomly for each feature and then
selects the best split among these \cite{Geurts-etal-06}. Experiments are
conducted using cross-validation under identical conditions for each algorithm.

\begin{figure}
[ptb]
\begin{center}
\includegraphics[
height=2.2831in,
width=4.3275in
]%
{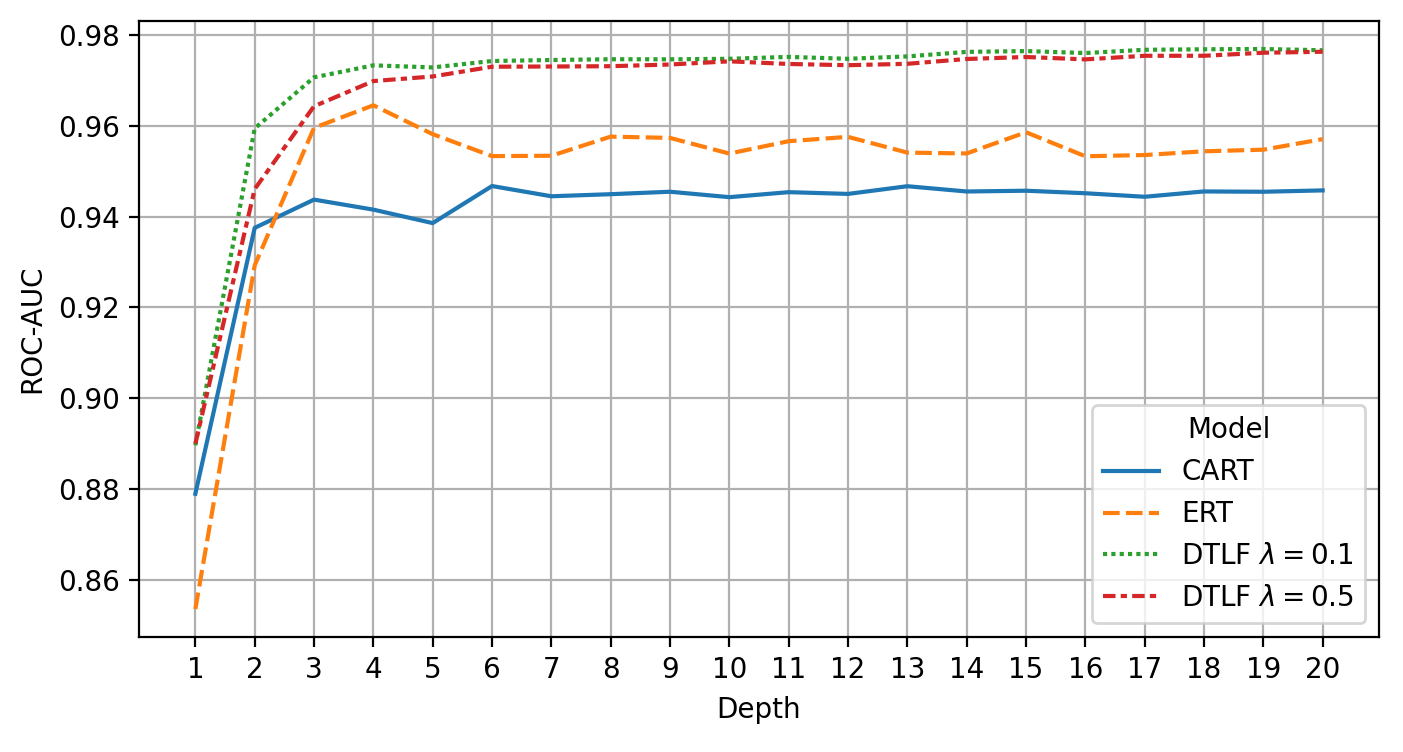}%
\caption{The classification accuracy (ROC-AUC) as a function of the tree depth
for different models trained on the BC dataset}%
\label{f:bc}%
\end{center}
\end{figure}

For classification tasks, which include anomaly detection with labeled data,
experiments are conducted on a real dataset. As an example, Fig. \ref{f:bc}
shows a comparison of ROC-AUC scores for trees constructed using classical
algorithms (CART, ERT) and trees constructed using the new algorithm DTLF
(with two different regularization parameters), depending on tree depth, for
the Breast cancer dataset. From Fig. \ref{f:bc}, it can be seen that when
using regularization, the proposed algorithm allows building models with
accuracy comparable to the maximum accuracy of the best classical tree (in
this case, the ERT algorithm) when the tree depth is only two. As the depth
increases, the proposed algorithm allows building even more accurate models,
while the accuracy of classical trees is limited to a lower value.%

ROC-AUC scores obtained using two methods (CART and ERT) and DTLF by two
values of the regularization parameter $\lambda=0.1$ and $\lambda=0.5$ (see
(\ref{regul_1})) trained on different datasets are shown in Table
\ref{t:classific}. The best results in all tables are shown in bold. It can be
seen from Table \ref{t:classific} that DTLF outperforms CART and ERT almost
for all datasets. One can again see from Table \ref{t:classific} that DTLF
with hyperparameter $\lambda=0.1$ provides better results than the model with
$\lambda=0.5$.%

\begin{table}[tbp] \centering
\caption{Accuracy measures (ROC-AUC) for the classification tasks}%
\begin{tabular}
[c]{ccccc}\hline
Dataset & CART & ERT & DTLF, $\lambda=0.1$ & DTLF, $\lambda=0.5$\\ \hline
Breast cancer & $0.942$ & $0.956$ & $\mathbf{0.974}$ & $0.973$\\
Car & $0.981$ & $0.981$ & $\mathbf{0.993}$ & $0.993$\\
Dermatology & $0.577$ & $0.563$ & $0.647$ & $\mathbf{0.674}$\\
Diabetic retinopathy & $0.636$ & $0.665$ & $\mathbf{0.671}$ & $0.660$\\
Ecoli & $0.764$ & $0.775$ & $\mathbf{0.871}$ & $0.868$\\
Eeg eyes & $0.854$ & $0.629$ & $\mathbf{0.893}$ & $0.889$\\
Haberman & $0.606$ & $0.654$ & $0.649$ & $\mathbf{0.658}$\\
Ionosphere & $0.894$ & $0.913$ & $0.925$ & $\mathbf{0.926}$\\
Seeds & $0.952$ & $0.948$ & $\mathbf{0.967}$ & $0.962$\\
Seismic & $0.578$ & $0.607$ & $0.691$ & $\mathbf{0.706}$\\
Soybean & $\mathbf{0.976}$ & $0.974$ & $0.968$ & $0.963$\\
Teaching assistant & $0.672$ & $\mathbf{0.673}$ & $0.657$ & $0.653$\\
Tic-tac-toe & $0.950$ & $0.951$ & $\mathbf{0.970}$ & $0.962$\\
Website phishing & $0.947$ & $0.945$ & $0.963$ & $\mathbf{0.965}$\\
Wholesale customers & $0.499$ & $\mathbf{0.503}$ & $0.498$ & $0.485$\\ \hline
\end{tabular}
\label{t:classific}%
\end{table}%

\begin{figure}
[ptb]
\begin{center}
\includegraphics[
height=3.32in,
width=5.1837in
]%
{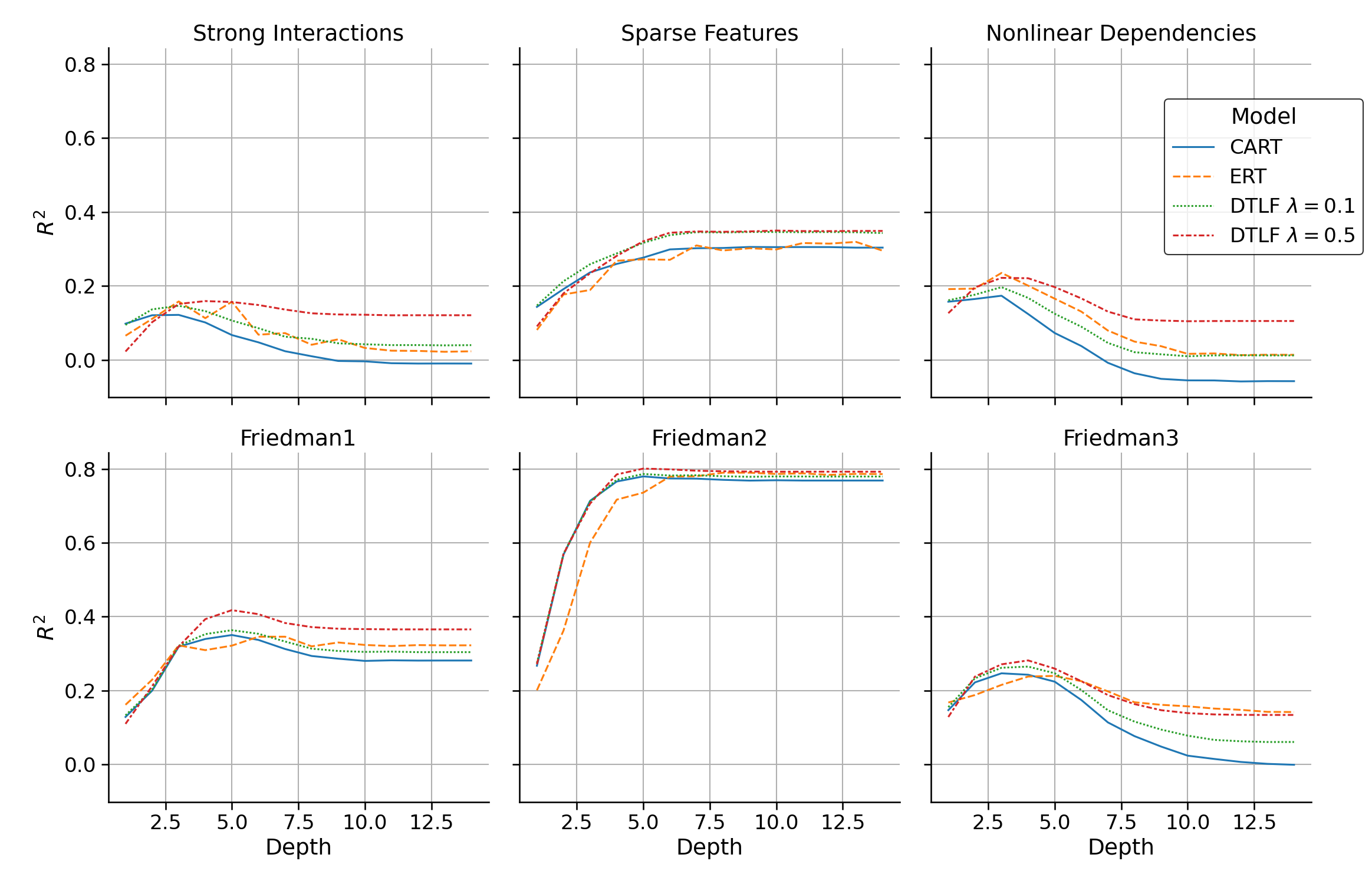}%
\caption{The regression accuracy ($R^{2}$) as a function of the tree depth for
CART and DTLF trained on six datasets}%
\label{f:regression_6}%
\end{center}
\end{figure}

For analyzing the algorithm in solving regression tasks, we first study the
models with synthetic datasets (\emph{Strong Interactions}, \emph{Sparse
Features}, \emph{Nonlinear Dependencies}, \emph{Friedman1}, \emph{Friedman2},
\emph{Friedman3}), for which 400 examples with 5 features are generated. The
target variable (event time) has a Weibull distribution with parameter $k=5$
and an expectation defined by the given rule (functional dependence on
features). In this series of experiments, all events are observed (no
censoring), so $R^{2}$ is used as the accuracy metric. Fig.
\ref{f:regression_6} shows the dependence of accuracy ($R^{2}$, higher is
better) on the tree depth for the CART algorithm, ERT and DTLF with two
regularization options: $\lambda=0.1$ and $\lambda=0.5$, where all
hyperparameters of the algorithms are the same (except for regularization,
which is only present in DTLF). The quadratic error is used as the loss
function for DTLF. As can be seen from Fig. \ref{f:regression_6}, the accuracy
of trees constructed using DTLF is higher than that of the CART algorithm and
ERT at the same tree depth and other hyperparameters (minimum number of
examples in a leaf is 3, minimum number of examples for splitting is 6). It
can be noted that regularization significantly affects generalization ability,
as clearly seen in several cases: \emph{Nonlinear Dependencies},
\emph{Friedman1}, \emph{Friedman3}, where larger regularization parameter
values significantly improve accuracy. Moreover, the optimal regularization
parameter values differ for different datasets. For example, for the
\emph{Sparse Features} dataset, which has the same number of training examples
as the other datasets, larger regularization values lead to accuracy
comparable to the CART algorithm, while smaller regularization values improve accuracy.%

\begin{figure}
[ptb]
\begin{center}
\includegraphics[
height=2.1179in,
width=4.4365in
]%
{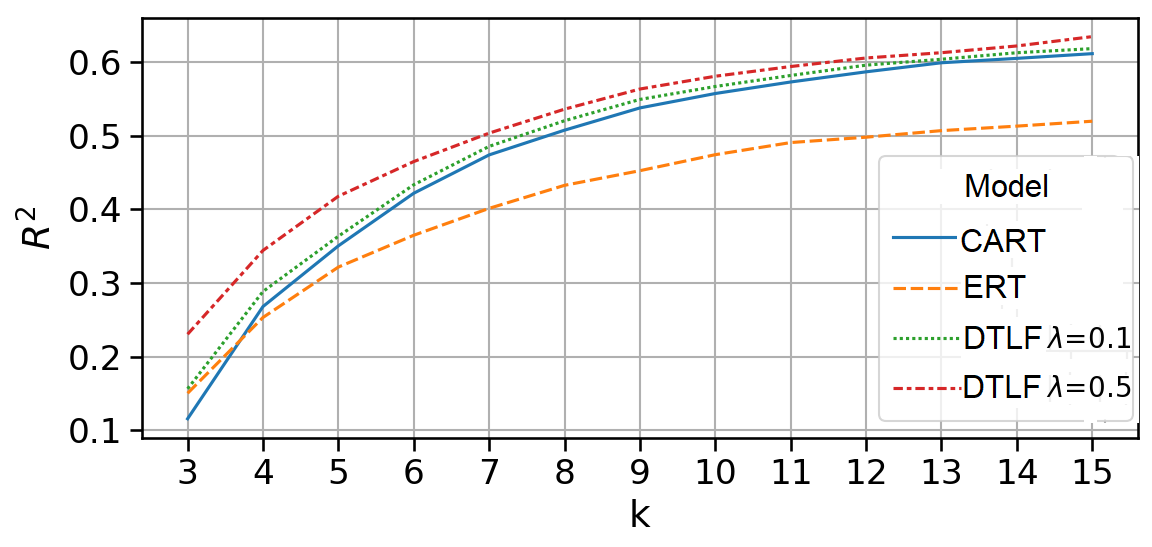}%
\caption{The regression accuracy ($R^{2}$) as a function of the parameter $k$
for CART, ERT, and DTLF trained on the Friedman1 dataset}%
\label{f:friedman1_by_k}%
\end{center}
\end{figure}

Consider in more detail the task with one dataset, \emph{Friedman1}, with
different values of the Weibull distribution parameter $k$. For a fixed depth
of $5$, the dependence of accuracy on $k$ is shown in Fig.
\ref{f:friedman1_by_k}. The larger the value of $k$, the smaller the variance
of the predicted value. It can be seen that in this case, with larger
regularization parameter values, DTLF achieves higher accuracy for all
considered values of $k$. The difference in accuracy is more significant for
smaller $k$ and higher variance, indicating that regularization helps avoid
overfitting. It is also worth noting that for $k=3$, the CART algorithm with a
deterministic tree construction algorithm is less accurate than ERT, where
splitting rules are partially determined randomly, while the proposed model
does not exhibit such degradation in accuracy despite being deterministic.%

\begin{figure}
[ptb]
\begin{center}
\includegraphics[
height=2.0358in,
width=4.4797in
]%
{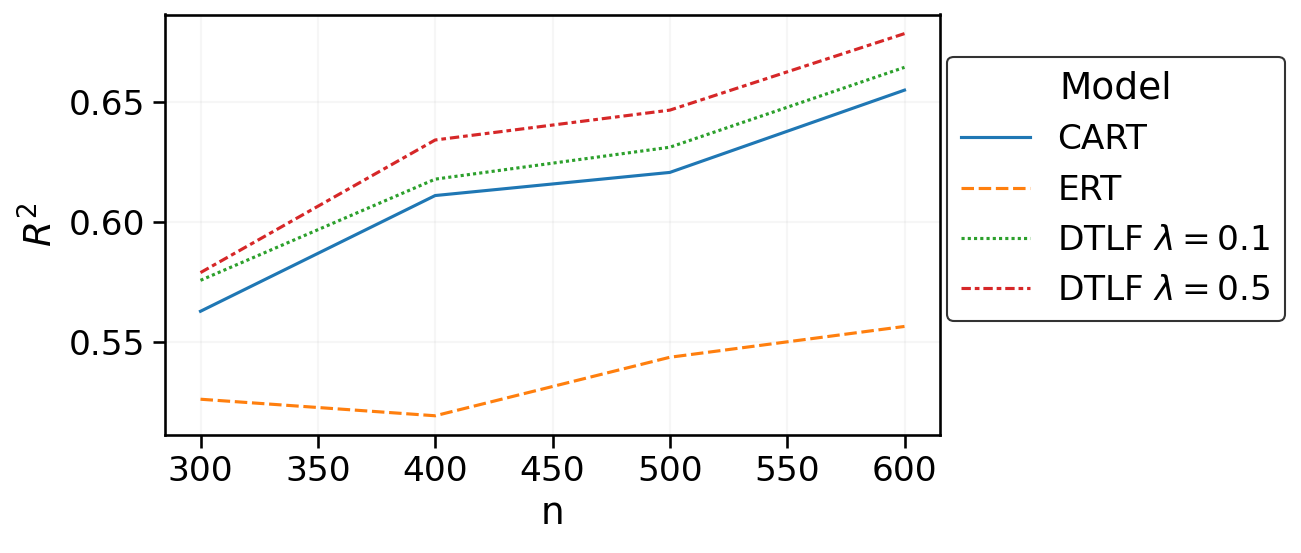}%
\caption{The regression accuracy ($R^{2}$) dependent on the sample size $n$
for CART and DTLF trained on the Friedman1 dataset}%
\label{f:friedman1_by_n}%
\end{center}
\end{figure}

Note that the superiority of DTLF persists even with low variance of the
predicted value ($k=15$), including with increasing data size. This is
demonstrated in Fig. \ref{f:friedman1_by_n}, where for a fixed value of $k=15$
and the depth $5$, the dependence of accuracy on the sample size is shown for
each method (CART on the left, then DTLF with $\lambda=0.1,0.5$). As can be
seen from Fig. \ref{f:friedman1_by_n}, when using DTLF with a regularization
parameter of $\lambda=0.5$, the accuracy remains higher than that of the
classical algorithm as the sample size increases.%

$R^{2}$ measures obtained using two methods (CART and ERT) and DTLF by four
values $0.01$, $0.1$, $0.5$, $1$ of the regularization parameter $\lambda$
trained on different regression datasets are shown in Table \ref{t:regression}%
. It can be seen from Table \ref{t:regression} that DTLF with $\lambda=1$
outperforms other models. However, this outperformance is not significant.
However, if we look at all results obtained by using DTLF for all parameters
$\lambda$, the we can say that DTLF outperforms CART and ERT for most datasets.%

\begin{table}[tbp] \centering
\caption{Accuracy measures ($R^{2}$) for the regression tasks}%
\begin{tabular}
[c]{ccccccc}\hline
& CART & ERT & \multicolumn{4}{c}{DTLF}\\
Dataset &  &  & $\lambda=0.01$ & $\lambda=0.1$ & $\lambda=0.5$ & $\lambda
=1$\\ \hline
Airfoil & $\mathbf{0.837}$ & $0.828$ & $0.815$ & $0.820$ & $0.821$ & $0.721$\\
Boston & $0.739$ & $0.680$ & $0.758$ & $0.750$ & $0.772$ & $\mathbf{0.776}$\\
Concrete & $0.813$ & $0.784$ & $0.808$ & $0.812$ & $\mathbf{0.820}$ &
$0.815$\\
Diabetes & $-0.101$ & $-0.110$ & $0.030$ & $0.080$ & $0.145$ & $\mathbf{0.204}%
$\\
Wine & $0.037$ & $0.011$ & $0.150$ & $0.177$ & $0.229$ & $\mathbf{0.265}$\\
Yacht & $0.991$ & $0.990$ & $0.992$ & $\mathbf{0.992}$ & $0.991$ &
$0.986$\\ \hline
\end{tabular}
\label{t:regression}%
\end{table}%

\begin{figure}
[ptb]
\begin{center}
\includegraphics[
height=3.3762in,
width=5.6922in
]%
{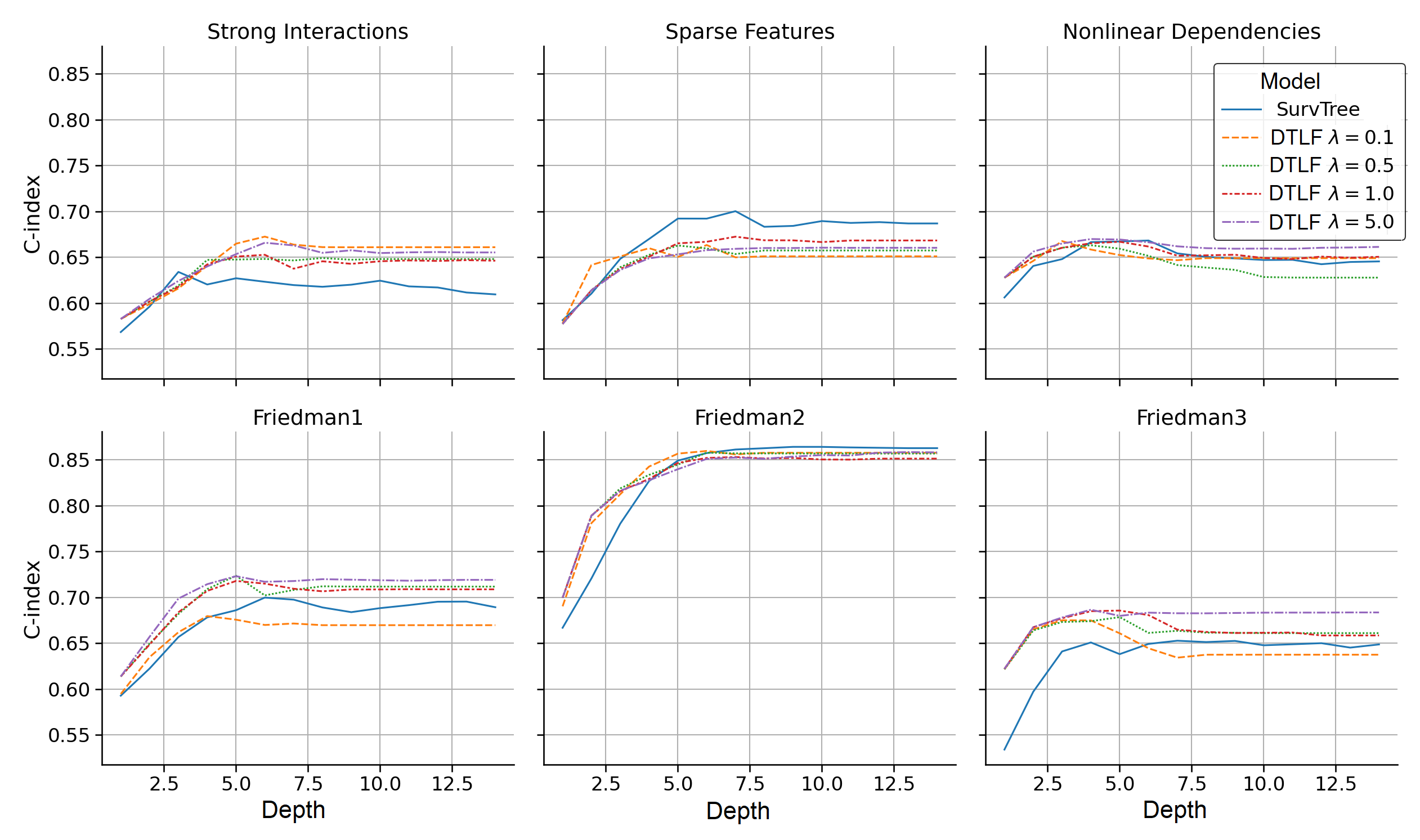}%
\caption{The survival regression accuracy (C-index) as a function of the tree
depth for the survival tree (SurvTree) and DTLF trained on six datasets}%
\label{f:survival_reg}%
\end{center}
\end{figure}

For predictive analytics tasks with censored data, experiments are conducted
on synthetic datasets. A total of 20\% of randomly selected events are
censored. The concordance index (C-index) is used to evaluate accuracy. Fig.
\ref{f:survival_reg} shows a comparison of the accuracy of the existing
decision tree construction algorithms for survival analysis tasks, SurvTree
\cite{Bou-Hamad-etal-2011,Gordon-Olshen-1985}, and DTLF for different
regularization parameter values, depending on the tree depth. From Fig.
\ref{f:survival_reg}, it can be seen that for most datasets, the maximum
regularization parameter value $\lambda=5$ achieves the highest accuracy. For
the \emph{Sparse Features} and \emph{Friedman2} datasets, the SurvTree
algorithm achieves higher accuracy.%

The next question is how different methods for selecting estimates in leaves
of the tree in DTLF impact on the accuracy. We consider the method of refining
logits in leaves and the Kaplan-Meier estimate based on all examples in the
leaf. Moreover, we also compare different initialization methods:

\begin{enumerate}
\item initializing logits via the Kaplan-Meier estimate denoted as
\textquotedblleft'with initialization\textquotedblright;

\item initializing logits with zero values denoted as \textquotedblleft%
'without initialization\textquotedblright.
\end{enumerate}

\begin{figure}
[ptb]
\begin{center}
\includegraphics[
height=3.2837in,
width=5.8237in
]%
{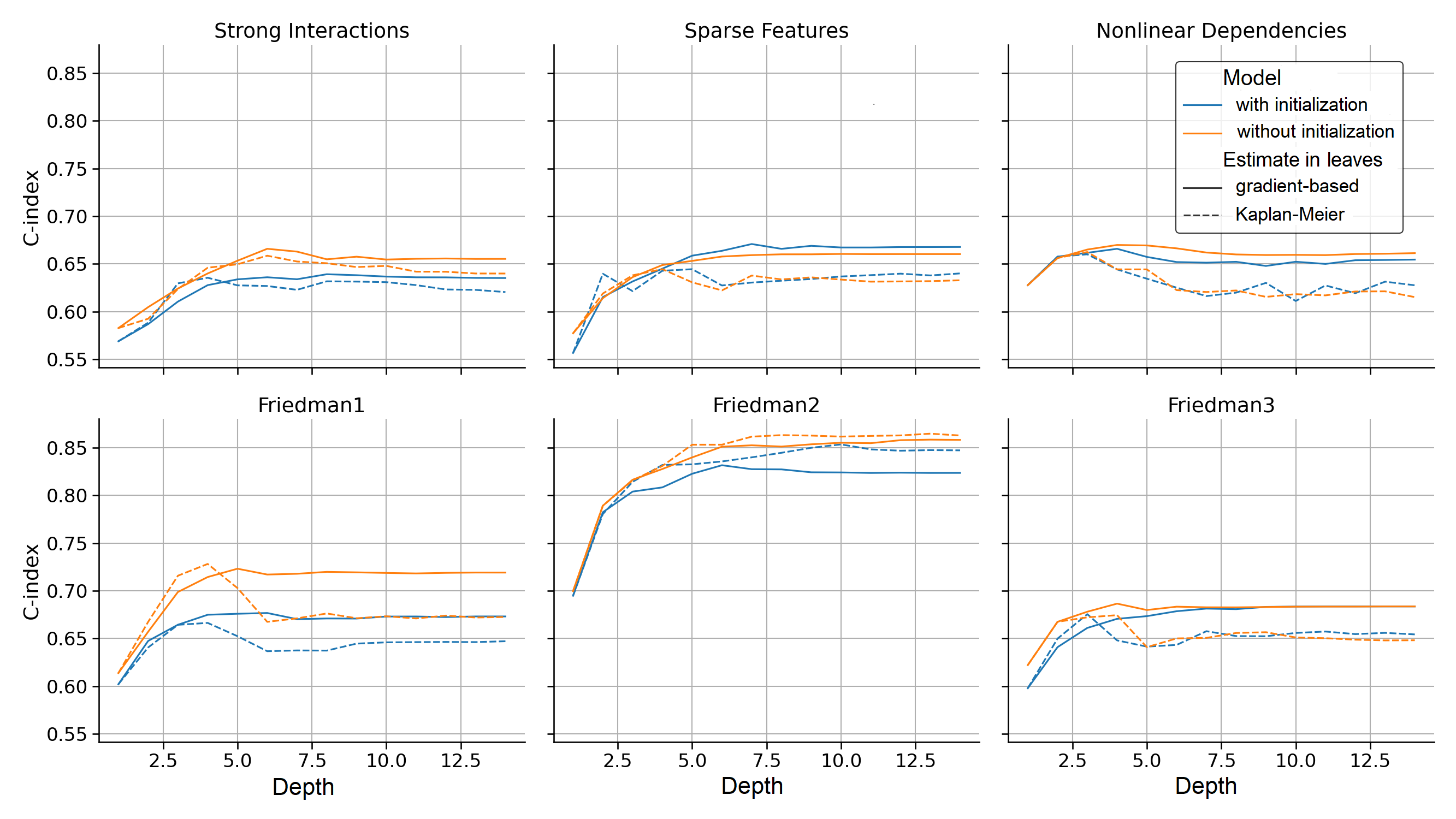}%
\caption{The survival regression accuracy (C-index) as a function of the tree
depth for different training conditions}%
\label{f:km_vs_ge}%
\end{center}
\end{figure}

The comparison results are shown in Fig. \ref{f:km_vs_ge}. It can be noted
that, first, when using Kaplan-Meier estimates in the leaves, accuracy
decreases starting from a certain tree depth, which can be explained by the
reduction in the number of training examples in the leaves. Second, using
initialization often does not provide advantages compared to zero initialization.%

\section{Conclusion}

We have introduced a novel gradient-based method for constructing decision
trees, referred to as DTLF (Decision Tree with Loss Function). This method
optimizes arbitrary differentiable loss functions, addressing key limitations
of traditional decision tree algorithms such as CART and ERT. By leveraging
the first and second derivatives of the loss function, DTLF enables more
accurate and flexible modeling, particularly in complex tasks such as
classification, regression, and survival analysis.

Extensive numerical experiments on both synthetic and real-world datasets have
demonstrated the superiority of DTLF over traditional methods. In
classification tasks, DTLF achieved higher ROC-AUC scores, particularly with
regularization, which helped mitigate overfitting and improve generalization.
For regression tasks, DTLF consistently outperformed CART and ERT in terms of
$R^{2}$ scores. In survival analysis, DTLF showed competitive performance in
terms of the concordance index (C-index).

The proposed method is implemented in a publicly available software package,
providing a practical tool for researchers and practitioners to apply and
extend the approach to their own datasets and tasks.

It should be noted that the proposed method has been developed specifically
for decision trees and not for ensembles of decision trees. However, it can be
extended to random forests and gradient boosting, which represents a promising
direction for further research.

An interesting problem is to explore how the proposed method can be applied to
tasks beyond classification, regression, and survival analysis, such as
clustering, anomaly detection, interpretation, and feature selection. Each of
these tasks is characterized by a specific loss function, and the proposed
approach opens the door to solving multitask machine learning problems, which
is another exciting direction for future research.

Finally, the approach supports the integration of decision trees with neural
networks, creating hybrid models capable of handling structured data (e.g.,
sequences, graphs) while maintaining the interpretability of tree-based
models. The combination of decision trees and neural networks is a key area
for further exploration.

\bibliographystyle{unsrt}
\bibliography{Boosting,Classif_bib,Deep_Forest,Explain,MYBIB,Survival_analysis}

\appendix

\section*{Appendix}

\section{Software Implementing the Proposed Method}

To implement the proposed new algorithms for constructing decision trees based
on a gradient approach, software packages were developed in Python version
3.11. To improve performance, extensions were implemented in Cython version
3.0.11 (transpiled to C and C++), allowing the construction of trees in
accordance with the proposed gradient algorithm. Loss functions were
implemented, along with the ability to specify arbitrary loss functions,
including functions written in Python, enabling the composition of decision
trees and neural networks implemented using the PyTorch library. The source
code of the packages is publicly available at: \url{https://github.com/NTAILab/gradient_growing_trees}.

The initial implementation of the algorithm was done without using extensions
and is available in the package as
\emph{gradient\_growing\_trees.naive.FullGTreeRegressorMultiOutput}. This
model always represents a full tree of a given depth; if no training examples
fall into a node, zero values are recorded in it. Additionally, the model's
predictions are determined by the values of the corresponding leaves, so, for
example, when implementing classification models, post-processing (e.g.,
\emph{softmax}) is required to obtain probability vectors. To search for
threshold values, either an exhaustive search or a search over randomly
generated values is used, with the number of such values determined by the
\emph{n\_guess} parameter. Prediction is implemented as a tree traversal,
calculating refinements to the predictions at each point, using point indices
to determine the refinements. The loss functions described above, including
quadratic error, cross-entropy, and a generalized loss function based on
log-likelihood, were implemented. To define new loss functions, a subclass of
\emph{BaseLossGradHess} must be defined, with a function that takes the true
values and current estimates in the tree node as input.

The classical CART algorithm cannot be directly extended by changing the
splitting criterion, as the developed algorithm requires computing values not
only in leaf nodes but also in internal nodes, based not on training example
labels but on the derivatives of the loss function and the values obtained in
the parent node. Moreover, the quality of the split cannot be characterized by
a measure such as impurity (as in CART), since only the values of the loss
function derivatives are known. Therefore, the implementation includes:

\begin{itemize}
\item a tree construction algorithm using depth-first traversal, which does
not use impurity as a stopping criterion;

\item an algorithm for searching for locally optimal splits;

\item criteria evaluating the quality of potential splits and computing
refined values in the nodes.
\end{itemize}

The splitting criterion determines the refined value in the current node,
calculates the loss function derivatives for each training example falling
into the node, and then computes the total derivative values. The split search
algorithm iterates over features and sorts the indices of examples falling
into the leaf to order the features in ascending order. Thus, to compute the
loss function, it is sufficient to add one observation to the left and remove
one observation from the right. The loss function depends on the accumulated
sums of derivatives to the left and right of the threshold, and therefore its
value can be updated in $O(1)$ operations. Thus, the optimal split can be
found in $O(m)$ time, where $m$ is the number of examples in the node.

Two types of splitting criteria were implemented. The first type, denoted in
the code as \emph{GradientCriterion}, evaluates derivatives for each example
separately and does not require additional memory to store derivative values.
The main limitation of such criteria is the need to repeatedly compute
derivatives (for each feature, derivatives are recalculated). The second type,
denoted as \emph{BatchGradientCriterion}, evaluates derivatives once by
storing the first and second derivatives for each training example in
pre-allocated memory. This implementation is significantly more
computationally efficient for computationally complex loss functions, such as
those with high-dimensional predicted values or when combining decision trees
with neural networks. An external loss function can be used via
\emph{BatchArbitraryLoss} (of the second type). To compute derivatives, the
external function is passed the training example labels, the indices of
examples in the leaf, the current predictions in the node, and buffers to
store the first and second derivatives for the passed indices. A global
interpreter lock is used to call the external function, which may slow down
execution; however, since the loss function is computed for each node rather
than each example, this aspect does not significantly impact performance.

From a practical perspective, when using the developed package, an external
loss function can be defined as follows (in this example, a quadratic loss
function is implemented):
\begin{verbatim}
def batch_custom_loss(ys, sample_ids, cur_value, g_out, h_out):
    g_out = np.asarray(g_out)
    h_out = np.asarray(h_out)
    sample_ids = np.asarray(sample_ids)
    y = np.asarray(ys)[sample_ids]
    g_out[sample_ids] = 2.0 * (np.asarray(cur_value)[np.newaxis] - y)
    h_out[sample_ids] = 2.0

model = GradientGrowingTreeRegressor(
    lam_2=0.1,
    max_depth=8,
    criterion=batch_custom_loss
)
\end{verbatim}

An example of interaction with the PyTorch library to obtain derivative
values, assuming the loss function module (denoted below as
\emph{torch\_nn\_module}) is implemented in PyTorch and takes as input the
training labels and current predictions in the leaf for each example:
\begin{verbatim}
def torch_based_loss(ys, sample_ids, cur_value, g_out, h_out):
    g_out = np.asarray(g_out)
    h_out = np.asarray(h_out)
    sample_ids = np.asarray(sample_ids)
    values_tensor = torch.tensor(
        np.asarray(cur_value)
    ).unsqueeze(0).repeat(sample_ids.shape[0], 1).requires_grad_(True)
    ys_tensor = torch.as_tensor(np.asarray(ys)[sample_ids])
    loss = torch_nn_module(ys_tensor, values_tensor)
    grads, *_ = torch.autograd.grad(loss, values_tensor)
    g_out[sample_ids] = grads.numpy()
    h_out[sample_ids] = 1.e-3

model = GradientGrowingTreeRegressor(
    lam_2=0.1,
    max_depth=8,
    criterion=torch_based_loss
)
\end{verbatim}

As seen from the above code snippets, interaction with the developed package
is straightforward and allows for rapid implementation of new models by
specifying derivative loss functions.
\end{document}